\documentclass[11pt]{article}

\usepackage[]{acl2021}
\usepackage{times}
\usepackage{latexsym}

\usepackage{xcolor}

\usepackage{bm}
\usepackage{amsthm,amsmath,amssymb}
\usepackage{mathrsfs}
\usepackage{epsfig}
\usepackage{graphicx}
\usepackage{booktabs}
\usepackage{verbatim}
\usepackage{multirow}
\usepackage{algorithm}
\usepackage{algorithmicx}
\usepackage{algpseudocode}

\usepackage{tabularx}
\usepackage{algorithm}
\usepackage{algorithmicx}
\usepackage{algpseudocode}
\usepackage{graphicx}
\usepackage{subcaption} 
\usepackage[normalem]{ulem}
\usepackage{cleveref}

\crefformat{section}{\S#2#1#3} % see manual of cleveref, section 8.2.1
\crefformat{subsection}{\S#2#1#3}
\crefformat{subsubsection}{\S#2#1#3}

\usepackage{microtype}

\algnewcommand\algorithmicinput{\textbf{Input:}}
\algnewcommand\INPUT{\item[\algorithmicinput]}
\algnewcommand\algorithmicoutput{\textbf{Output:}}
\algnewcommand\OUTPUT{\item[\algorithmicoutput]}

\title{Investigating Data Variance in Evaluations \\ of Automatic Machine Translation Metrics}

\author{Jiannan Xiang$^{\heartsuit}$\thanks{~~Work done while J. Xiang was an intern at Tencent AI Lab.}\ , Huayang Li$^{\spadesuit}$, Yahui Liu$^\bigstar$, Lemao Liu$^\clubsuit$, \\ \textbf{Guoping Huang$^\clubsuit$, Defu Lian$^\diamondsuit$, Shuming Shi$^\clubsuit$} \\
  $^\heartsuit$Carnegie Mellon University, $^\spadesuit$Nara Institute of Science and Technology \\ $^\diamondsuit$University of Science and Techology of China \\
  $^\bigstar$University of Trento, Italy, $^\clubsuit$Tencent AI Lab\\
  \texttt{\small jiannanx@cs.cmu.edu, li.huayang.lh6@is.naist.jp} \\ \texttt{\small yahui.liu@unitn.it, lemaoliu@gmail.com,} \texttt{\small \{donkeyhuang, shumingshi\}@tencent.com} \\
  }

\begin{document}
\maketitle
\begin{abstract}
%Automatic metrics serve an important role in machine translation, thus it is critical to ensure that the metric evaluation protocols are reliable. 
Current practices in metric evaluation mainly focus on one single dataset from a targeted domain, \emph{e.g.}, News domain in each year's WMT Metrics Shared Task. In this paper, we qualitatively and quantitatively show that the performances of metrics are sensitive to data even when the data is from the same domain, i.e., the ranking of metrics varies when the evaluation is conducted on different datasets from the same domain. Then this paper further investigates 
two potential hypotheses, \emph{i.e.}, insignificant data points and the deviation of Independent and Identically Distributed (i.i.d) assumption, which may take responsibility for the issue of data variance. In conclusion, our findings suggest that when evaluating automatic translation metrics, researchers should take data variance into account and be cautious to claim the result on a single dataset even from the same domain, because it may leads to inconsistent results with most of other datasets.
\end{abstract}

\section{Introduction}
Assessing the quality of machine translation (MT) systems is always crucial to promote MT research~\cite{marie-etal-2021-scientific}. Since it is costly and time-consuming for human graders to evaluate machine translation (MT) systems, designing automatic metrics for MT has drawn booming attention during the past decades, and many metrics such as BLEU~\citep{papineni-etal-2002-bleu} and TER~\citep{snover2006study} have been proposed 
consequently.

Generally, it is non-trivial to measure automatic metrics.  Conference Machine Translation (WMT)~\citep{ma-etal-2019-results,ma-etal-2018-results,bojar-etal-2017-results,bojar2016results, bojar2016ten} thereby holds the Metric Shared Task to evaluate the performance of automatic metrics. In each year, WMT organizers collect datasets consisting of many MT outputs annotated with human judgments, and automatic metrics are evaluated on the dataset in terms of their correlations to human judgments.
{In recent two years, WMT started to evaluate the metrics on the datasets from different domains and found that the evaluation results are not domain robust~\cite{freitag2021results, mathur2020results}. However, it is still unclear whether the evaluation results for the datasets from the same domain are variant or not. Since each year the dataset from the News domain was mainly involved for evaluation, 
%Over the past ten years, the official evaluation reports independently analyzed the results for the dataset of each year. To the best of our knowledge, there are no studies to examine the evaluation results for different datasets from the same domain and make a more systematic analysis.~\footnote{In recent two years, WMT started to evaluate the metrics on the datasets from different domains and found that the evaluation results are not domain robust~\cite{freitag2021results, mathur2020results}. However, it is unclear whether the evaluation results for the datasets from the same domain are variant or not.}
%However, involved metrics are different year by year, thus it is difficult to comparatively ana yze the evaluation results among different years.
%Therefore, 
some key questions remain unknown: are the evaluation results consistent across different years?
%\textcolor{red}{What influences will the use of different datasets have on the evaluation results?} 
Are the results on each dataset reliable?}

%To answer these questions, we evaluate ten popular metrics on all available datasets in the past ten years and empirically investigate the fluctuation of metric evaluation results.
One may simply summarize the existing results from the official evaluation reports of the past years and answer the above questions accordingly. %
However, the existing results use Pearson's correlation for evaluation which suffers from sensitivity to outlier data points as argued by~\citet{mathur2020tangled}.
%\yahui{are all existing summarizing methods based on Pearson's correlation? If not, the ``However" is a bit strange.}
Besides, involved metrics in the evaluation are different year by year, thus it is difficult to directly compare the results among different years.
To this end, in this work, we firstly re-evaluate ten popular metrics on many datasets from News domain between 2010 and 2019, with the Error Number evaluation method~\citep{mathur2020tangled}.  We then empirically investigate the fluctuation of metric evaluation results. Surprisingly, our experiments show that the evaluation result is sensitive to the choice of datasets, which suggests that the results on some datasets may not be reliable (\S 3).

% \begin{table*}[htp!]
% \small
% \centering
% \begin{tabularx}{\linewidth}{cccc}
%     \toprule
%     Dataset & Size & System Number & Link  \\
%     \midrule
%     Newssyscombtest2010 &  2,034 &31& \url{http://www.statmt.org/wmt10/results.html}\\
%     Newssyscombtest2011 &  2,000 &26& \url{http://www.statmt.org/wmt11/results.html} \\
%     Newstest2012  & 3,003 &16& \url{http://www.statmt.org/wmt12/results.html} \\
%     Newstest2013  & 3,000 &23& \url{http://www.statmt.org/wmt13/results.html} \\
%     Newstest2014  & 3,003 &13& \url{http://www.statmt.org/wmt14/results.html} \\
%     Newstest2015  & 2,169 &13& \url{http://www.statmt.org/wmt15/results.html} \\
%     Newstest2016  & 2,999 &10& \url{http://www.statmt.org/wmt16/results.html} \\
%     Newstest2017  & 3,004 &11& \url{http://www.statmt.org/wmt17/results.html} \\
%     Newstest2018  & 2,998 &16& \url{http://www.statmt.org/wmt18/results.html} \\
%     Newstest2019  & 2,000 &16& \url{http://www.statmt.org/wmt19/results.html}\\
%     \bottomrule
% \end{tabularx}
% \caption{\label{dataset} The data statistics for German-English language pair.}
% \end{table*}

\begin{table}[htp!]
\small
\centering
\begin{tabularx}{\linewidth}{ccc}
    \toprule
    Dataset & Size & System Number  \\
    \midrule
    Newssyscombtest2010 &  2,034 &31\\
    Newssyscombtest2011 &  2,000 &26 \\
    Newstest2012  & 3,003 &16\\
    Newstest2013  & 3,000 &23\\
    Newstest2014  & 3,003 &13\\
    Newstest2015  & 2,169 &13\\
    Newstest2016  & 2,999 &10\\
    Newstest2017  & 3,004 &11\\
    Newstest2018  & 2,998 &16\\
    Newstest2019  & 2,000 &16\\
    \bottomrule
\end{tabularx}
\caption{\label{dataset} The data statistics for German-English language pair.}
\end{table}

Then we investigate two potential hypotheses about the emergence of data variance, i.e., the insignificant data points (\S 4.1) and deviation of Independent and Identically Distributed (i.i.d) assumption (\S 4.2). First, we show that the data variance issue is substantially alleviated when the insignificant data points are removed. To further understand the variance that cannot be alleviated by the first hypothesis, we design a simple method to measure the distributional differences between datasets, and hypothesize that the deviation of the i.i.d assumption may contribute to the variance. %\lemao{Maybe this byproduct is not good enough.} \textcolor{blue}{\sout{As a byproduct, from the re-evaluation results, we are able to figure out some unreliable datasets which lead to inconsistent results with others.}}
For future metric evaluation, we recommend WMT community pay attention to the potential issue of data variance when conducting evaluations. 
% issue of data variance and be cautious to report the evaluation results on such an unreliable dataset.

%Overall, we find that current metrics evaluation results obtained from one dataset may not be reliable, since the results change when evaluation is conducted on different datasets. For future metrics evaluation, we recommend the inclusion of all the available datasets instead of using a single dataset.

\begin{table}[htp!]
\small
\centering
\begin{tabularx}{\linewidth}{ccc}
    \toprule
    Metrics & Features & Average Type \\
    \midrule
    BLEU & n-grams & macro \\
    WER & Levenshtein distance &  macro \\
    TER & edit distance & macro \\
    PER & edit distance & macro \\
    chrF & character n-grams & micro \\
    chrF+ & character n-grams & micro \\
    BEER & char. n-grams, trees & micro \\
    CharacTER & char. edit distance & micro  \\
    BERTScore & neural representations & micro \\
    MoverScore & neural representations & micro \\
    \bottomrule
\end{tabularx}
\caption{\label{metrics}Features for the metrics we use in the paper. Note that we implement PER by ourselves.}
\end{table}

\section{Experiment Settings}
\subsection{Datasets and evaluation metrics}
% Since 2008, WMT conducted the Metrics Shared Task\footnote{Homepage: \url{http://www.statmt.org/wmt19/metrics-task.html}} 
% each year. Before 2016, the task collected Relative Ranking (RR)~\citep{bojar2016results} as the gold human assessments. To get RR, human annotators were provided with source sentence, target reference, and five distinct MT output translations, and were required to rank the five translations from best to worse with ties allowed. We use Expected Wins~\citep{callisonburch-EtAl:2012:WMT} as the score of an MT system.
% Starting from \citet{bojar-etal-2017-results}, the task switched to Direct Assessment (DA) \citep{graham2013continuous}, where annotators were asked to rate the adequacy of a set of translations compared to the corresponding source/reference sentence. The performance of an MT system is presented by the mean of the standardised score of all its translations. 
We collect the testing datasets and the human assessments of the WMT Metrics Task from 2010 to 2019, where the datasets are from the same News domain. %\textcolor{blue}{\sout{Due to space limitations, in this paper, we only report the results on the German-English language pair, but the conclusions are consistent with all other language pairs. Results of other language pairs can be found in the appendix.} 
In this work, we mainly conduct experiments on the De$\Rightarrow$En task and more details about datasets are shown in Table~\ref{dataset}. However, as shown in \cref{sec:vdd}, our conclusions are consistent on other translation tasks, such as Ru$\Rightarrow$En.

Since participating metrics in the WMT Metrics Task varied over years, we collect ten most popular metrics and re-evaluate them on all ten datasets to study their performance.%, because our goal is not to find the best metric among all existing metrics.
These metrics are summarized as follows:
BLEU~\citep{papineni-etal-2002-bleu}, WER~\citep{morris2004and}, PER~\citep{tillmann1997accelerated}, TER~\citep{snover2006study}, chrF~\citep{popovic-2015-chrf}, chrF+~\citep{popovic2017chrf++}, BEER~\citep{stanojevic-simaan-2014-fitting}, CharacTER~\citep{wang2016character}, BERTScore~\citep{zhang2019bertscore}, and MoverScore~\citep{zhao2019moverscore}. {The first 4 metrics are in system-level (i.e., macro) while others are in sentence-level (i.e., micro), as shown in Table~\ref{metrics}. Since extending sentence-level metrics to system-level is more natural \cite{zhang2019bertscore}, we only compare them on the system-level.}  

\begin{table*}[]
\small
\centering
\setlength{\tabcolsep}{4pt}
\begin{tabular}{lcccccccccc}\toprule
\multirow{3}{*}{\textbf{Metric}} & \multicolumn{10}{c}{\textbf{Dataset}}   \\ 
\cmidrule(r){2-11}
                     & 2010    & 2011   & 2012    & 2013    & 2014    & 2015   & 2016    & 2017    & 2018   & 2019   \\ \midrule
BERTScore                & \textbf{1} / 24.4  & \textbf{1} / 37.1 & \textbf{2} / 28.9  & \textbf{1} / 10.6  & \textbf{2} / 20.4  & \textbf{1} / 14.7 & \textbf{1} / 14.5  & \textbf{6} / 24.6  & \textbf{2} / 15.3 & \textbf{3} / 37.0 \\
CharacTER                & \textbf{6} / 27.6  & \textbf{1} / 37.1 & \textbf{1} / 24.2  & \textbf{6} / 18.0  & \textbf{1} / 17.3  & \textbf{1} / 14.7 & \textbf{3} / 17.6  & \textbf{1} / 20.8  & \textbf{1} / 14.4 & \textbf{4} / 38.2 \\
MoverScore               & \textbf{2} / 25.2  & \textbf{4} / 39.3 & \textbf{2} / 28.8  & \textbf{2} / 11.7  & \textbf{2} / 20.3  & \textbf{1} / 14.7 & \textbf{2} / 16.0  & \textbf{5} / 23.9  & \textbf{2} / 15.4 & \textbf{1} / 36.6 \\
chrF                     & \textbf{3} / 26.7  & \textbf{1} / 37.8 & \textbf{4} / 29.7  & \textbf{2} / 12.1  & \textbf{2} / 20.8  & \textbf{4} / 17.7 & \textbf{4} / 18.9  & \textbf{2} / 22.9  & \textbf{2} / 15.3 & \textbf{1} / 37.0 \\
BEER                     & \textbf{3} / 26.3  & \textbf{5} / 45.3 & \textbf{5} / 33.5  & \textbf{4} / 13.4  & \textbf{6} / 25.0  & \textbf{5} / 19.0 & \textbf{5} / 19.5  & \textbf{2} / 23.2  & \textbf{2} / 15.2 & \textbf{6} / 38.4 \\
chrF+                    & \textbf{3} / 26.9  & \textbf{5} / 45.8 & \textbf{6} / 35.1  & \textbf{4} / 13.8  & \textbf{7} / 26.4  & \textbf{6} / 19.2 & \textbf{5} / 20.2  & \textbf{2} / 23.3  & \textbf{2} / 15.2 & \textbf{4} / 37.7 \\
BLEU                     & \textbf{8} / 32.3  & \textbf{8} / 58.3 & \textbf{8} / 42.3  & \textbf{7} / 20.9  & \textbf{8} / 29.3  & \textbf{7} / 23.1 & \textbf{8} / 21.2  & \textbf{7} / 26.3  & \textbf{9} / 18.1 & \textbf{7} / 41.3 \\
WER                      & \textbf{7} / 31.7  & \textbf{7} / 57.7 & \textbf{7} / 40.8  & \textbf{8} / 23.4  & \textbf{9} / 32.3  & \textbf{7} / 22.9 & \textbf{5} / 19.7  & \textbf{8} / 27.2  & \textbf{7} / 17.0 & \textbf{7} / 40.9 \\
TER                      & \textbf{9} / 35.0  & \textbf{9} / 61.2 & \textbf{9} / 43.9  & \textbf{9} / 24.7  & \textbf{10} / 36.2 & \textbf{7} / 22.7 & \textbf{8} / 20.9  & \textbf{10} / 28.6 & \textbf{7} / 17.2 & \textbf{9} / 43.0 \\
PER                      & \textbf{10} / 38.6 & \textbf{9} / 61.7 & \textbf{10} / 48.0 & \textbf{10} / 26.9 & \textbf{5} / 23.8  & \textbf{7} / 22.8 & \textbf{10} / 28.2 & \textbf{8} / 27.6  & \textbf{9} / 18.4 & \textbf{9} / 43.5\\ \bottomrule
\end{tabular}
\caption{\label{table-1} Metric evaluation results on De$\Rightarrow$En datasets from 2010 to 2019. The tuple "\textbf{R} / E" shows the performance of a metric, where \textbf{R} denotes Significant Ranking (\S 2.3) among all metrics and E denotes the Error Rate (Error Number divided by the total number of system pairs).}
% Ranking clusters are identified by grouping metrics together according to which metrics significantly outperform all others in lower-ranking clusters.}
\vspace{-1em}
\end{table*}

%\subsection{System Comparison Approach}
For each system pair, metrics or humans give a comparison result about whether one system is better than another. Following \citet{graham-etal-2014-randomized}, we use statistical significance tests to detect if the difference in scores (metrics or humans) between two systems is significant. Specifically, for RR scores, we use the bootstrap method~\citep{koehn-2004-statistical}. For DA scores, we apply the Wilcoxon rank-sum test. For macro-average metrics, \emph{i.e.}, BLEU, WER, PER, and TER, we use the bootstrap method~\citep{koehn-2004-statistical}. For other micro-average metrics, we use the paired t-test method.

% \begin{figure*}[t]
% \centering
% \subfigure[120 system pairs]{\label{sign_heatmap}
% \begin{minipage}[t]{0.5\linewidth}
% \centering
% \includegraphics[width=0.85\linewidth]{acl-ijcnlp2021-templates/sign_heatmap.pdf}
% %\caption{fig1}
% \end{minipage}%
% }%
% \subfigure[10k system pairs]{\label{sh_heatmap}
% \begin{minipage}[t]{0.5\linewidth}
% \centering
% \includegraphics[width=0.85\linewidth]{acl-ijcnlp2021-templates/sign_hybrid_heatmap.pdf}
% %\caption{fig2}
% \end{minipage}%
% }%
% \centering
% \vspace{-2ex}
% \caption{Pairwise metrics comparison results on newstest2019 German-English dataset. A green cells indicates that the error number of the metric in the row is significantly less than that of the metric in the corresponding column.}
% \label{sign}
% \end{figure*}

\subsection{Measuring Automatic Metrics}\label{evaluation}
The previous WMT Metrics Tasks used Pearson's $r$ to measure the ability of a metric to evaluate MT systems. However, as pointed out by \citet{mathur2020tangled}, Pearson's $r$ is unstable for a small sample size and sensitive to outlier systems. Besides, in practice, metric scores are always used to compare pairs of MT systems\footnote{Unless otherwise specified, a system always denotes MT system in our work, rather than an evaluation metric.}. Thus following \citet{mathur2020tangled}, we measure an automatic metric with the number of errors made by the metric when comparing system pairs. Error Number can be considered as an absolute view of measuring a metric. 
%In addition, we provide a relative view to measure a metric, Significant Ranking, which is based on Error Number. 
%Measurement of automatic metrics are presented in detail in the following. 

\paragraph{Error Number} 
% Following~\citet{mathur2020tangled}, we measure the performance of a metric with the number of errors made by the metric.
Following~\citet{mathur2020tangled}, we measure the performance of a metric by its consistency with humans.
Specifically, each metric or human can judge whether a system performs better compared to another system (details of system comparison process are presented in the appendix), and the error number is the number of contrary cases between the results of metric and human. %(i.e., counting all available MT system pairs whether they are consistent with gold human judgments or not). 
%There are three types of errors: 
%Type \uppercase\expandafter{\romannumeral1} error occurs when human judge one system to be better than the other while the metric judges there is no significant difference. Type \uppercase\expandafter{\romannumeral2} error occurs when metric difference is significant while the difference of human judgments is insignificant. The third type of error occurs when both differences are significant but are opposite to each other. %The error number for a metric is sum of the error numbers for all three types. 
As mentioned by~\citet{graham2016achieving}, when the number of compared MT systems are too small on a dataset, differences among different metrics may be insignificant. Thus, the results of the metric evaluation can be highly inconclusive. We indeed observe similar results in our experimental setting. Therefore, we use the hybrid super-sampling method~\citep{graham2016achieving} to create a large number of hybrid system pairs: on each dataset, we synthesize 142 systems in total, which form 10K system pairs. Finally, we calculate the error number of each metric on all 10K system pairs.

\subsection{Measuring Data Variance}

\paragraph{Significant Ranking}
Based on the measurement of error number, a qualitative approach to know whether those metrics perform consistently on different datasets is to evaluate the variance of their rankings. To make the ranking more reliable, we propose a \textit{significant ranking} method, which conducts significant test when sorting the error numbers of metrics. For example, in Table~\ref{table-1}, the significant ranking of all metrics on 2010 dataset is ``1, 6, 2, 3, 3, 3, 8, 7, 9, 10" where chrF, BEER and chrF+ are with the same relative ranking of 3. This is because they are not significantly different, although their absolute error numbers are slightly different.
%of BLEU may be smaller than TER, the ranks of BLEU and TER are regarded as the same, if the difference between them are not statistically significant. 
% In metric evaluation, we are also interested in pairwise ranking between two metrics. Although the error numbers can indicate which metric is better than the other, such a pairwise ranking may not be statistically significant. 
We employ the bootstrap re-sampling method~\cite{koehn-2004-statistical} to test if the number of errors of one metric is significantly less than the others. For the bootstrap method, we repeat re-sampling 1000 times and set the p-value to 0.05 for all the significance tests.

\paragraph{Disagreement Number}
In addition, we also propose a method to quantitatively measure the variance between two datasets $\mathcal{D}$ and $D'$, namely, \textit{disagreement number}. Specifically, we construct a set $\mathcal{S}_{D}$ by collecting all pairwise metrics that one is significantly better than the other on dataset $\mathcal{D}$. Then to measure the mismatch between $\mathcal{D}$ and $\mathcal{D}^\prime$, we count the {\bf disagreement number} between the pairwise metrics in $\mathcal{S}_{\mathcal{D}}$ and that in $\mathcal{S}_{\mathcal{D}^\prime}$. For example, disagreement number plus one, if BLEU is significantly better than TER on $\mathcal{D}$ and worse than TER on $\mathcal{D}^\prime$. Although this number is linear to Kendall's Tau~\cite{kendall1938new}, it is able to show more informative difference between two overall rankings. For example, two metrics with totally different ranks may just have a slight difference on disagreement number. 
%\sout{between two significant overall rankings regarding two datasets. %~\footnote{For example, for 2010 and 2011 datasets, xxx (please explain how to calculate $\mathcal{S}_{D}$, the disagreement between both datasets and Kendall's Tau, respectively).}
As a result, we employ disagreement number rather than Kendall's Tau to show the quantitative difference between two overall rankings more intuitively for more detailed analysis.  
It is worth mentioning that the disagreement number is at most 45 in our setting where there are 10 metrics in total.  

% \paragraph{Hybrid Super-Sampling} Empirically, we find that the number of system pairs is generally too small for metrics to show significant difference between each other, which leads to highly inconclusive results. Therefore, we leverage the hybrid super-sampling method~\citep{graham2016achieving} to create a large number of hybrid system pairs. Finally, on each dataset we have 142 systems in total, which form 10k system pairs.
% \paragraph{Parameters} For bootstrap method, we repeat resampling a 1000 times. We set the p-value as 0.05 for all the statistical significance test.

% \paragraph{System Pair Comparison} For each system pair, metrics or humans give a comparison result about whether one system is better than another. Following \citet{graham-etal-2014-randomized}, we use statistical significance tests to detect if the difference in scores (metrics or humans) between two systems is significant. Specifically, for RR scores, we use the bootstrap method~\citep{koehn-2004-statistical}. For DA scores, we apply the Wilcoxon rank-sum test. For macro-average metrics, \emph{i.e.}, BLEU, WER, PER, and TER, we use the bootstrap method~\citep{koehn-2004-statistical}. For other micro-average metrics, we use the paired t-test method.

\section{Data Variance in Metric Evaluations}\label{realdata}
% In this section, we firstly demonstrate that metric evaluation results are sensitive to the choice of datasets, then we analyze its potential reasons and finally we propose a simple method to verify whether a dataset is sufficient.\lemao{Please do not mention sufficient data.}

% Annual WMT Metrics Shared Task has been held for over ten years, and the task collected and annotated the Newstest dataset for metric evaluation every year. However, although the datasets from previous years are available, the task only reports the evaluation results on the single dataset of that year. In this paper, we would like to explore whether the MT metric evaluation results on one dataset is reliable or not and how dataset variance affects the results. 

%which suggests that results obtained on one single dataset may not be reliable. We detail our experiments as follows.

\begin{figure}[t]
\begin{center}
   \includegraphics[width=1\linewidth]{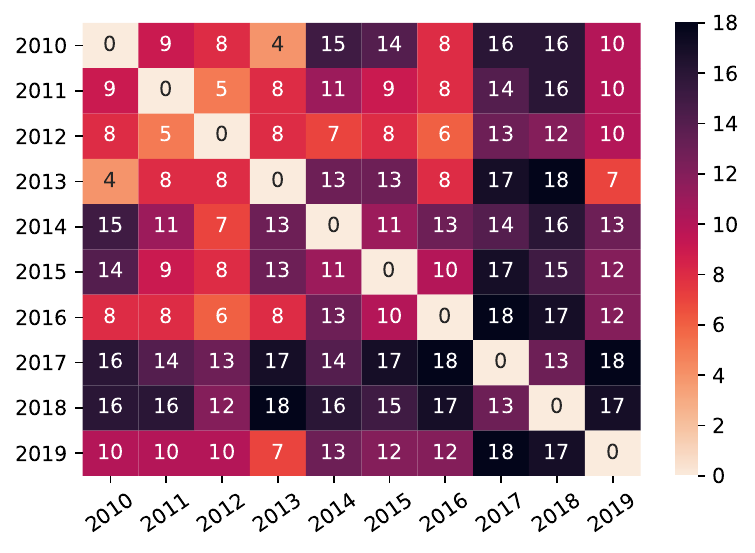}
\end{center}
   \caption{The heatmap for the disagreement numbers between every two datasets on  De$\Rightarrow$En task.}
\label{disagree_heatmap}
\vspace{-1em}
\end{figure}

%Hybrid super-sampling is a kind of way to manually create pseudo data from limited real data. Previous WMT metrics tasks conducted metrics evaluation on the dataset of that year, and have also adopted the hybrid method to overcome the problem of limited data and systems. However, we would like to ask the question: Can pseudo data replace real data? Given that all previous WMT datasets are available, is it reasonable to ignore them and only focus on one year's dataset because creating enough pseudo data can reduce or even eliminate the uncertainty of the evaluation results? To answer the question, we conduct several experiments, and the results show that the metrics ranking is sensitive to the dataset, which suggests that the evaluation results on one single dataset may not be reliable. We detail our experiments in the following. 

\subsection{Variance of Different Datasets\label{sec:vdd}}
%We conduct experiments on all 10 datasets. We have 10 metrics, which can form 45 metric pairs. Therefore, on each dataset, we have 45 pairwise metric ranking. For each metric pair, we have 10 rankings on 10 different datasets. %We then examine whether the metrics comparison results agree with each other between different datasets.

% \begin{table*}[]
% %\small
% \centering
% %\setlength{\tabcolsep}{4pt}
% \begin{tabular}{lcccccccccc}\toprule
%  \textbf{Dataset}  & 2010    & 2011   & 2012    & 2013    & 2014    & 2015   & 2016    & 2017    & 2018   & 2019   \\ \midrule
%  Mean           & 9.28    & 3.11   & 2.45    & 3.95    & 4.63    & 7.93   & \textbf{14.71}   & 8.79    & 5.84   & \underline{13.66}  \\
%  Variance          & 8.20     & 2.28   & 1.13    & 3.97    & 4.03    & 5.27   & \textbf{22.53}   & 9.27    & 5.15   & \underline{20.60}   \\      
% \bottomrule
% \end{tabular}
% \caption{The statistics (mean and variance) of disagreement numbers between split subsets. \label{tab:sufficient}}
% \end{table*}

We conduct experiments on all 10 datasets. We have 10 metrics, which can form 45 metric pairs. On each dataset, for each metric, we calculate its error number (described in Section~\ref{evaluation}).
%, whose definition is described before. 
In addition, we perform a statistical significance test for each metric pairs in terms of both error numbers, from which we can obtain a ranking result among 10 metrics accordingly.

Table~\ref{table-1} illustrates the error numbers and ranks on 10 datasets. 
It shows that the ranks are always variant along with different datasets. 
For example, on the dataset of 2011, the error rate of MoveScore is larger than chrF (39.3 v.s. 37.8), and the former ranks 4 while the latter ranks 1. However, it is opposite on the dataset of 2015, where MoveScore ranks 1 with an error rate of 14.7 while chrF ranks 4 with an error rate of 17.7. {As shown in Table \ref{supplementary}, we observe a similar trend on the Ru$\Rightarrow$En task.} 
%where MoveScore ranks 1 and chrF ranks 4 with the error rates of 14.7 and 17.7, respectively.

% To better measure the variance between two datasets $D$ and $D'$, we further study the mismatches of overall results between different datasets. Specifically, we collect all significant pairwise rankings to constitute a set $\mathcal{S}_{D}$ for each dataset $D$. Then to measure the mismatch between $D$ and $D'$, we count the disagreement number between the pairwise rankings in $\mathcal{S}_{D}$ and that in $\mathcal{S}_{D'}$. Note that this number is linear to Kendall's Tau~\cite{kendall1938new} between two significant overall rankings regarding two datasets.%~\footnote{For example, for 2010 and 2011 datasets, xxx (please explain how to calculate $\mathcal{S}_{D}$, the disagreement between both datasets and Kendall's Tau, respectively).}
%  Disagreement number shows the difference between two overall rankings more intuitively, so we report it rather than Kendall's Tau. 
 
As shown in Figure~\ref{disagree_heatmap}, there is a high inconsistency between the results of different datasets and none of the dataset pairs achieve zero disagreements. The difference between the datasets in 2010 and 2013 is the smallest (i.e., only 4 disagreed metric pairs). However, most of the disagreement numbers are larger than 10 (the maximum achieves 18). Moreover, datasets from 2017 to 2019 not only have a high disagreement number with datasets of early years, but also have high variances among themselves. This finding is a little surprising, because in our sense the quality of WMT's datasets must be improved year by year.

\begin{table}[]
\centering
\resizebox{1.0\columnwidth}{!}{
\setlength{\tabcolsep}{4pt}
\begin{tabular}{lccccc}\toprule
\multirow{3}{*}{\textbf{Metric}} & \multicolumn{5}{c}{\textbf{Dataset}}   \\ 
\cmidrule(r){2-6}
                         & 2015   & 2016    & 2017    & 2018   & 2019   \\ \midrule
BERTScore   & \textbf{2} / 18.0 & \textbf{3} / 16.7  & \textbf{2} / 32.1  & \textbf{3} / 24.5 & \textbf{1} / 35.8 \\
CharacTER   & \textbf{6} / 20.5 & \textbf{6} / 19.4  & \textbf{1} / 30.4  & \textbf{1} / 22.3 & \textbf{4} / 37.2 \\
MoverScore  & \textbf{1} / 14.9 & \textbf{1} / 15.1  & \textbf{2} / 31.4  & \textbf{3} / 24.5 & \textbf{1} / 36.1 \\
chrF        & \textbf{3} / 18.7 & \textbf{2} / 15.7  & \textbf{2} / 31.9  & \textbf{2} / 24.0 & \textbf{4} / 37.1 \\
BEER        & \textbf{3} / 19.0 & \textbf{3} / 17.0  & \textbf{5} / 33.2  & \textbf{3} / 24.3 & \textbf{9} / 39.6 \\
chrF+       & \textbf{5} / 19.7 & \textbf{5} / 17.6  & \textbf{5} / 33.3  & \textbf{3} / 24.5 & \textbf{4} / 36.9 \\
BLEU        & \textbf{10} / 27.9 & \textbf{7} / 21.0  & \textbf{9} / 34.8  & \textbf{8} / 25.0 & \textbf{9} / 39.8 \\
WER         & \textbf{8} / 23.4 & \textbf{7} / 21.5  & \textbf{5} / 33.2  & \textbf{8} / 25.0 & \textbf{8} / 37.9 \\
TER         & \textbf{8} / 23.4 & \textbf{10} / 23.3  & \textbf{9} / 34.5 & \textbf{10} / 25.8 & \textbf{4} / 36.9 \\
PER         & \textbf{6} / 21.1 & \textbf{7} / 21.5 & \textbf{7} / 34.0  & \textbf{2} / 23.4 & \textbf{1} / 35.7 \\ \bottomrule
\end{tabular}}
\caption{Metric evaluation results on Ru$\Rightarrow$En datasets from 2015 - 2019.}\label{supplementary}
\end{table}

\section{Potential Reasons for Data Variance}

Many factors may contribute to the data variance issue, but lots of them are difficult to be evaluated, such as the personal preferences of humans. In this section, we propose to analyze two potential factors that can be quantitatively evaluated.

\subsection{Insignificant Data Points}

\begin{figure}[t]
\begin{center}
   \includegraphics[width=1\linewidth]{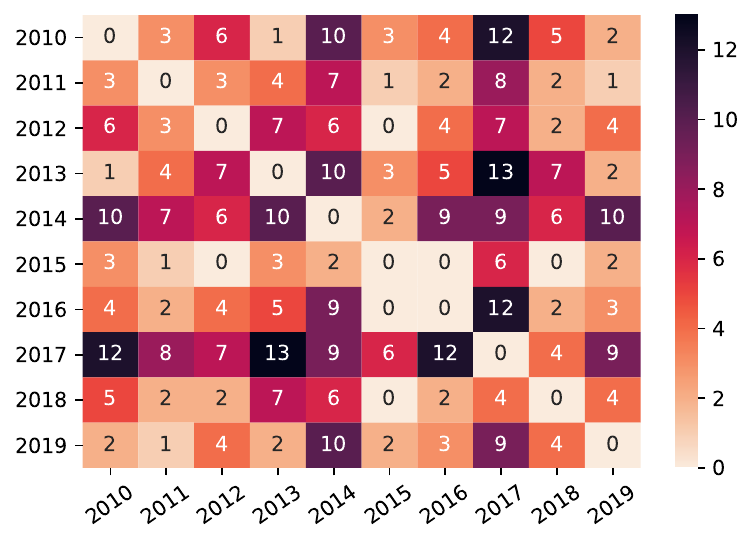}
\end{center}
   \caption{The heatmap for the disagreement numbers between every two datasets on De$\Rightarrow$En task. Insignificant  system pairs according to human assessments are removed.}
\label{significant_disagree_heatmap}
\vspace{-1em}
\end{figure}

Intuitively, if the translations $H_{A}$ from system $A$ are much better than those $H_B$ from system $B$ in translation quality according to human evaluation, then it is easy to judge the better system even for a weak automatic metric. In contrast, if $H_{A}$ is similar to $H_{B}$ 
in translation quality, it is typically difficult to judge the better system even for a good metric. This motivates us that such an insignificant data point $\langle H_A, H_B\rangle$ may take responsibility for the data variance issue. 

To validate the above intuition, 
we remove the system pairs that are not significantly different according to human evaluation, and compute the disagreement number between any two datasets again. 
The experimental results are shown in Figure \ref{significant_disagree_heatmap}. We can see the disagreement number decreases greatly comparing to the results in Figure \ref{disagree_heatmap}. In the previous experiment, most of the disagreement numbers are greater than 10, while in the new experiment most of them are less than 5, and some of them even achieve 0, such as the number between 2012 and 2015, which means the ranks of metrics are exactly the same on those datasets. 
The results indicate that part of the data variance issue can be explained by system pairs that humans think are not significantly different.

However, After the removal of insignificant data points, some disagreement number are still high, e.g., the number between 2013 and 2017 is 13. It demonstrates that there are still some other underlying problems that give rise to the data variance phenomenon. In addition, the datasets for both 2014 and 2017 do not agree well with others. This indicates that we should be cautious to report overall results on some datasets, e.g., 2014 and 2017.

\subsection{Deviation of I.I.D Assumption}
How to interpret the high variance on datasets, e.g., 2014 and 2017, remains to be an open question. In this section we try to give a hypothesis based on the i.i.d assumption. 
According to the principle of statistical sampling, if two samples are drawn from the same distribution, then a statement made on one sample is likely to hold on the other sample. Therefore, one hypothesis about the high variance may be that datasets from different years deviate i.i.d assumption. In fact, this may be true in our scenario because each dataset is generated by a set of translation systems but the set of systems is variant each year. 

To this end, we design an experiment to measure the extent to which each dataset is drawn from the same distribution during the past ten years.  
Since the standard method such as Kolmogorov-Smirnov test \cite{massey1951kolmogorov} is difficult to scale with respect to feature dimension, we employ adversarial validation to distinguish the difference between two datasets~\cite{pan2020adversarial}. Its basic idea is to formulate the i.i.d test as a classification problem and train a classifier between two datasets. If the classifier can accurately distinguish the data from different datasets, then the distributions of the two datasets are regarded as highly different. Since it is too slow to train classifiers for all pairs of datasets, we conduct experiments on three years from 2017 to 2019. More details are shown in appendix.  

The results on two kinds of datasets are shown in Table \ref{tab:acc}, where higher accuracy indicates clearer distributional differences between two datasets. Note that accuracy scores in main diagonal are got from two  subsets of each year via randomly splitting. As shown in Table \ref{tab:acc}, the distributional differences between MT datasets have been introduced by source sentences. After accompanied with the system outputs, the distributional differences are more severe between different years. This fact shows that some datasets in past ten years deviate the i.i.d assumption, which may be related to the inconsistency of metrics. 

\subsection{Suggestions}
According to those potential factors, we propose some suggestions to alleviate some potential issues for metric evaluation due to data variance in future. First, it would be better if pay more attention to those insignificant data points such that their manual annotations are as good as possible. Since it is costly to hire more annotators for data points, it would be possible to ask more annotators only for those insignificant data points. {Second, prior to reporting the evaluation results on a dataset, it would be helpful to pre-examine its data variance about the metric evaluation through some simulated approaches. For example, one possible approach is to randomly sample two subsets from the dataset without replacement and evaluate the consistency between the evaluation results on the two subsets, similar to the significant test in \cite{koehn-2004-statistical}.}
%Second, it would be helpful to construct a dataset with diverse domains and explicitly show the evaluation results for each subset with the same domain rather than a single evaluation result for the entire dataset. Generally, although inconsistent results from different domains are possible, however, the inconsistency in the same domain may be undesirable. 
% Second, it would be helpful to construct a dataset for evaluation where each sentence 
%Thus, showing the domain information could help researchers to better promote the datasets and metrics.

\section{Conclusion}

\begin{table}[]
    \centering
    \setlength{\tabcolsep}{3pt}{
    \subfloat[Src]{
    \begin{tabular}[t]{c|ccc}
    \hline
        & 17 & 18 & 19\\
        \hline
        17 & 50.4 & 52.8 & 65.8  \\
        18 & 52.8 & 51.4 & 67.5  \\
        19 & 65.8 & 67.5 & 50.9 \\
        \hline
    \end{tabular}}
    \quad
    \subfloat[Src+Output]{
        \begin{tabular}[t]{c|ccc}
        \hline
        & 17 & 18 & 19\\
        \hline
        17 &  51.4 & 75.3 & 80.2  \\
        18 & 75.3 & 55.6 & 79.2 \\
        19 & 80.2 & 79.2 & 52.2 \\
        \hline
    \end{tabular}}}
    \caption{The accuracy of classifiers. The higher value means two datasets deviate i.i.d assumption. We run the model with 5 different random seeds to calculate the average accuracy.}
    \label{tab:acc}
    \vspace{-1em}
\end{table}

Over the ten years between 2010 and 2019, the official evaluation reports of WMT Metrics Shared Task independently analyzed the results of that year. In this paper, we re-evaluate ten popular metrics on the datasets from the same domain in the ten years and comparatively analyze the evaluation results among different years together. We show the problem of conducting evaluations with only a single dataset. In addition, we analyze its potential reasons that the insignificant data points and deviation of i.i.d assumption may induce the issue of data variance. This fact suggests that future researches on evaluating automatic translation metrics should take data variance into account and be cautious to conclude the result on a single dataset. % that may lead to inconsistent results with other datasets. %inspired by the Law of Large Numbers. 
%We believe the conclusions made on all datasets are more credible than those obtained from one single dataset. 

\section*{Acknowledgement}
We would like to thank the anonymous reviewers and Markus Freitag for their constructive comments.

\bibliographystyle{acl_natbib}
\bibliography{acl2021}
\clearpage
\appendix

\section{Settings for Adversarial Validation}
To train the classifier, we need to construct a binary classification dataset first. Since the difference between distributions may come from both the source sentences and system outputs, we consider two types of classification datasets correspondingly. The first kind of dataset only considers the source information. Supposing that we want to compare the distributions of source sentences of MT datasets from year Y1 and Y2, we follow the three steps below to construct the classification dataset:

\begin{enumerate}
   \item We label the source sentences from Y1 and Y2 with 0 and 1, respectively;
   \item We split the data from Y1 and Y2 to train, dev, and test sets without overlap;
   \item We merge the data from Y1 and Y2 according to their split.
\end{enumerate}
For each pairs of MT datasets from year 2010 to 2019, we construct a classification dataset following the steps above. Besides the source information, we also construct  another kind of classification datasets that also consider the information of system outputs. The procedure to construct this kind of dataset is almost similar to the previous one, except that we concatenate each system outputs with its source sentences after the Step-2 is finished. In our experiments, we use mBERT \cite{devlin-etal-2019-bert, wolf-etal-2020-transformers} as the classifier, thus an unified structure can be used for the two kinds of datasets. 

\end{document}

% \maketitle
\appendix

% \section{Implementation Details}
% \subsection{Metrics Details}
% The details of the metrics used in the paper are provided in Table~\ref{metrics}.
% \begin{table*}[htp!]
% \small
% \centering
% \begin{tabularx}{\linewidth}{cccc}
%     \toprule
%     Metrics & Features & Average Type & Code \\
%     \midrule
%     BLEU & n-grams & macro & {\url{https://github.com/nltk/nltk}}~\citep{bird2009natural}\\
%     WER & Levenshtein distance &  macro & \url{https://github.com/jitsi/jiwer} \\
%     TER & edit distance & macro & \url{https://github.com/mjpost/sacrebleu}~\citep{post-2018-call} \\
%     PER & edit distance & macro & --- \\
%     chrF & character n-grams & micro & \url{https://github.com/m-popovic/chrF} \\
%     chrF+ & character n-grams & micro & \url{https://github.com/m-popovic/chrF} \\
%     BEER & char. n-grams, trees & micro & \url{https://github.com/stanojevic/beer} \\
%     CharacTER & char. edit distance & micro & \url{https://github.com/rwth-i6/CharacTER} \\
%     BERTScore & neural representations & micro & \url{https://github.com/Tiiiger/bert_score} \\
%     MoverScore & neural representations & micro & \url{https://github.com/AIPHES/emnlp19-moverscore}\\
%     \bottomrule
% \end{tabularx}
% \caption{\label{metrics}Features and code links for the metrics we use in the paper. Note that we implement PER by ourselves.}
% \end{table*}

%\section{Additional Experiments}
\section{Settings for Adversarial Validation}
To train the classifer, we need to construct a binary classification dataset first. Since the difference between distributions may come from both the source sentences and system outputs, we consider two types of classification datasets correspondingly. The first kind of dataset only considers the source information. Supposing that we want to compare the distributions of source sentences of MT datasets from year Y1 and Y2, we follow the three steps below to construct the classification dataset:

\begin{enumerate}
   \item We label the source sentences from Y1 and Y2 with 0 and 1, respectively;
   \item We split the data from Y1 and Y2 to train, dev, and test sets without overlap;
   \item We merge the data from Y1 and Y2 according to their split.
\end{enumerate}
For each pairs of MT datasets from year 2010 to 2019, we construct a classification dataset following the steps above. Besides the source information, we also construct  another kind of classification datasets that also consider the information of system outputs. The procedure to construct this kind of dataset is almost similar to the previous one, except that we concatenate each system outputs with its source sentences after the Step-2 is finished. In our experiments, we use mBERT \cite{devlin-etal-2019-bert, wolf-etal-2020-transformers} as the classifier, thus an unified structure can be used for the two kinds of datasets.